\documentclass[letterpaper,12pt]{article}
\usepackage{fullpage}
 
\usepackage{cite}
\usepackage{amsmath,amssymb,amsfonts}
\usepackage{algorithmic}
\usepackage{graphicx}
\usepackage{diagbox}
\usepackage{pdfpages}
\usepackage[square,sort,comma,numbers]{natbib}
\usepackage[margin=1in,letterpaper]{geometry} 
\usepackage{textcomp}
\usepackage{hyperref}
\hypersetup{
	pdftitle={Major Project Report Ankit Grover BTECH CCE 18930303158},
   colorlinks=true,
   citecolor=black,
   linkcolor=black,
   urlcolor=blue
}
\usepackage{verbatim}
\usepackage{pgfgantt}
\begin{document}
\begin{titlepage}

\begin{center}

\Large \textbf {Goodness of Pronunciation Pipelines for OOV Problem}\\[0.1in]
\normalsize{A  \\Major Project [CC1881] \\Report\\}
 \normalsize
       {Submitted in the partial fulfillment of the requirement for the award of \\ Bachelor of Technology \\in \\
Computer and Communication Engineering}\\[0.4 in]
        


\large Submitted by:\textbf{ \\ Ankit Grover \\ CCE 189303158}\\[0.3in]
\includegraphics[scale=0.8]{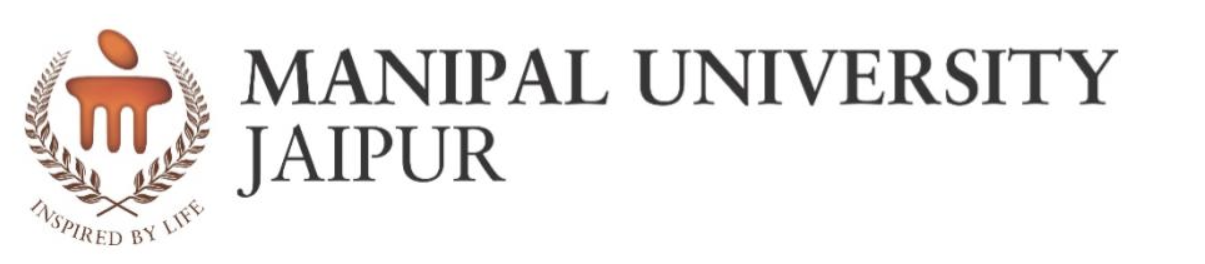}\\[0.1in]

\vspace{.3in}
Under the guidance of: \\
{\textbf{Dr. Vaishali Yadav }}\\

July 2022
\vfill
\rule{400pt}{2pt}
\large\textbf{{Department of Computer and Communication Engineering \\School of Computing and Information Technology\\Manipal University Jaipur}}\\
\normalsize VPO. Dehmi Kalan, Jaipur, Rajasthan, India – 303007

\end{center}

\end{titlepage}

\begin{center}
    \normalsize{Department of Computer and Communication Engineering \\School of Computing and Information Technology,Manipal University Jaipur \\VPO. Dehmi Kalan, Jaipur, Rajasthan, India – 303007}\\[0.5 in]
    
    \Large \textbf{STUDENT DECLARATION }\\[0.5 in]
    
    \normalsize{\textit{I hereby declare that this project \textbf{Goodness of Pronunciation Pipelines for OOV Problem} is my own work and that, to the best of my knowledge and belief, it contains no 
material previously published or written by another person nor material which has been 
accepted for the award of any other degree or diploma of the University or other Institute, 
except where due acknowledgements has been made in the text.
}}\\[1 in]

\large{Place: Jaipur, Rajasthan       \hfill         \textbf{Ankit Grover}}\\
\large{Date: 11/7/22      \hfill         189303158}\\
\large{\hfill B.Tech\ (CCE) 8th Semester
}\\

\end{center}

\pagebreak

\begin{center}
    \normalsize{Department of Computer and Communication Engineering \\School of Computing and Information Technology,Manipal University Jaipur \\VPO. Dehmi Kalan, Jaipur, Rajasthan, India – 303007}\\[0.5 in]
    
    \Large \textbf{CERTIFICATE FROM SUPERVISOR}\\[0.5 in]
    
    \normalsize{\textit{This is to certify that the work entitled “Goodness of Pronunciation Pipelines for OOV Problem” submitted by Ankit Grover 189303158 to Manipal University Jaipur for the award 
of the degree of Bachelor of Technology in Computer and Communication Engineering is a 
bonafide record of the work carried out by him/ her under my supervision and guidance from \textbf{Jan} to \textbf{July 2022.}
}}\\[1 in]

\large{\textbf{Dr. Vaishali Yadav (MUJ)} \hfill .\\Internal Supervisor \hfill .\\ Department of Computer and Communication Engineering \hfill .\\ Manipal University Jaipur. \hfill.}\\[1 in]
\large{Dr.Deepak Sinwar     \hfill         Prof. Vijaypal Singh Dhaka}\\
\large{\textbf{(Major Project Coordinator)}\hfill \textbf{(Head of Department)}
}\\

\end{center}

\pagebreak


\pagebreak 
\begin{center}

    \Large \textbf{ACKNOWLEDGEMENT}\\[0.5 in]
    
\large{I would like to thank my  advisor Dr. Vaishali Yadav for believing in my ideas and supporting them throughout the process. I would also like to thank Shubhankar Kamthankar for sharing his knowledge of linguistics which greatly helped me in understanding intuitions behind some of the decisions taken during implementations in my work. I would also like to thank, Jhansi Malela, Meenakshi, Snehal Ranjan who were ever present in helping me whenever I had questions or was stuck. I would also like to thank Aneesh Chavan, Shashwat Singh, Avani Gupta, Nishant Sachdeva who helped me to keep a positive outlook .Lastly, I would like to thank my parents, who have always supported me regardless of what stage of life I have been at. Their guidance has been priceless in my journey.}
\\[1 in]

\end{center}

\pagebreak

\begin{abstract}
In the following report we propose a  pipeline for Goodness of Pronunciation (GoP) computation solving OOV problem at Testing time using Vocab \& Lexicon expansion techniques and Lexicon Transducer re-construction.
The pipeline  uses different components of ASR 
system to quantify accent and automatically evaluate them as scores. We use the posteriors of 
an ASR model trained on native English speech, along with the phone level boundaries to obtain 
phone level pronunciation scores. We implemented methods 
to remove \emph{UNK} and \emph{SPN} phonemes arising in the GoP output due to absence of words in our Lexicon(Out of Vocabulary(OOV))  from a given utterance  by building three pipelines.The Online, Offline and Hybrid pipelines are proposed which take into account time and space complexity concerns with respect to WFST re-construction. The Online method is based per utterance, Offline method pre-incorporates a set of OOV words for a given data set and the Hybrid method combines the above two ideas to expand the lexicon as well work per utterance. We further provide utilities for  extracting DNN  posterior probabilities to Phone mapping , GoP scores as vectors for  each utterance , as well as Phone and Word boundaries to be used for future research purposes.

\end{abstract}

\pagebreak 

\tableofcontents
\pagebreak
\listoftables
\pagebreak
\listoffigures
\pagebreak

\

\section{INTRODUCTION} \label{intro}
\subsection{What is Speech Recognition}
Speech Recognition involves getting a correct transcript given Speech as an input.
Over the years many statistical methods have been proposed such as Dynamic Time Warping\cite{Permanasari_2019}, 
Linear Predictive Coding \cite{LPC}, Hidden Markov Models \cite{18626} and Deep Learning methods such as 
Deep Belief Networks \cite{huang2014research}, Recurrent Neural Networks \cite{huang2014research}, etc for speech
recognition. However, most of these methods do not consider accent information of
non-native speakers.
\\
\subsection{What is Accented Speech Recognition}
Accented Speech Recognition is the ability to correctly automatically get a transcription of the 
Speech which has accent influences. Models are often trained on data sets which lack such 
diversity. However, humans are diverse, which results in a huge set of accents. Each accent 
has distinct spectral characteristics such as difference in formants, prosody, pitch which make 
it distinguishable to the human ear and brain. However, unlike the human brain which can
correctly perceive accents, Machine Learning methods still fail to do so. This is the heart of 
the problem of Speech Recognition and is an important step in democratizing Speech 
Recognition technology. 
\
\subsection{Challenges associated with Accented Speech Recognition}
Despite the many advances in Speech Recognition, accented Speech Recognition is a very hot 
topic. People have proposed numerous methods for solving these problems. Accented Speech 
recognition has many challenges. In this section we describe the challenges that are present in 
the field. 
Many corporations approach the problem of Speech Recognition with the sole view of 
improving the various benchmarks, without tackling the actual problems. Their methods often 
lead to algorithmic bias and worsens performance of ASR systems \cite{markl2022language}, co-operation between 
communities is extremely important to mitigate such problems and approach Speech 
Recognition from a humanitarian perspective. A consequence of poorly performing ASR 
systems includes the phenomenon of people from small groups feeling excluded and many 
drop-out due to language barriers. ASR systems perform poorly in cases of code-switching 
\cite{vielzeuf2021e2e}, low-resource accented data, dialects \cite{aksenova2020algorithmic}, etc.
One of the biggest challenges for creating Accented Speech Recognition is the absence of 
high resource, diverse data-sets. 
Datasets for people with speech impairments is usually very less. These people face 
difficulties since the common ASR models fail to recognize their impaired Speech. Very few 
2
efforts have been made in this direction \cite{50460} \cite{kim2008dysarthric}, another problem associated with such 
databases is the language of collection of data \cite{garg2021towards}. It is important for data sets to have 
metadata such as age, gender, \cite{panayotov2015librispeech} other demographic information, if necessary, along with 
accented speech in multiple languages \cite{ardila2019common}. Very few data-sets contain Speech of non-native 
accents which is also very important for making the system robust and adapt to new accents.
More such data-sets are required \cite{aksenova2022accented}.
This makes training models on such data-sets robust to accent variations

\pagebreak

\section{EXPERIMENTAL SETUP}
We describe the Software and Hardware Requirements for our given experiments.

\subsection{Hardware Requirements}
We require a machine with at least 12 GB of RAM with an Nvidia GPU of at least 8GB 
VRAM for training of neural architectures. An Nvidia GPU of RTX 2060 is used in our setup. 
An Intel Core i7 with 8 cores and 16 threads is utilized. At least 1 TB of memory is required 
for downloading the required datasets and storing the results of the experiments. The 
Librispeech (960 hrs.) dataset was used in our experiments, because it has been shown to have 
more speaker variability, models trained on show better generalization capability .

\subsection{Software Requirements}

In our machine we require Python2.7, Python3, C++, Perl, shell, bash installed. It is 
recommended to install a package manager such as conda/pip, etc. for the same. In our 
experiments we utilize open-source tools and software such as Kaldi and Sequitur-G2P \cite{bisani2008joint}. 
We chose Kaldi toolkit \cite{povey2011kaldi} because of its ability to model complex hybrid ASR architectures 
based on Hidden Markov Models (HMM) and contains a whole host of other utilities for 
building an ASR system from scratch. It has a lot more utilities than comparable toolkits such 
as HTK, CMU Sphinx, etc. Kaldi is very powerful due to its speed which makes it suitable for 
deployment environments.

\pagebreak

\section{AUTOMATIC SPEECH RECOGNTION PARADIGMS}

There are different methods one can approach Speech Recognition in. This depends primarily 
on the type of problem we are tackling to solve and the availability of data. In many scenarios 
we require high interpretability despite high resourced data and might not need fine-grained 
analysis. Whereas in some cases we have low resourced data and require high accuracy and 
performance despite the bottleneck of less data. Thus, we can approach the Speech 
Recognition problem in 2 ways: Statistical/Hybrid Speech Recognition and End-to-End 
Speech Recognition.

The fundamental problem of Speech Recognition can be represented as :\
\begin{center}
$\hat{W}= \underset{W}{\mathrm{argmax}} P(W|X)$
\end{center}

where W, is the sequence of words(transcription) and X is the sequence of acoustic feature 
vectors (Observations).

\subsection{End-to-End Speech Recognition}

End-to-End models of Speech recognition involve training Neural Network based models,
without any extra modules. These Neural Network components can generalize different 
patterns present in the observation data and gain linguistic context without the presence of 
external Linguistic feedback in the form of Lexicons, Grammar Rules, etc. This makes Neural 
Networks highly desirable for usage in industrial as well as research purposes.
The figure below shows a general architecture of End-to-End Speech Recognition system. 
The input to the model is generally in the form of Mel frequency cepstrum coefficients or 
other features such as Linear Prediction Coefficients or Spectrogram directly.
The recognizer here is the neural network, which learns to model the speech from the features 
and outputs the result in the form of characters.

\begin{figure}
  \centering
  \includegraphics[]{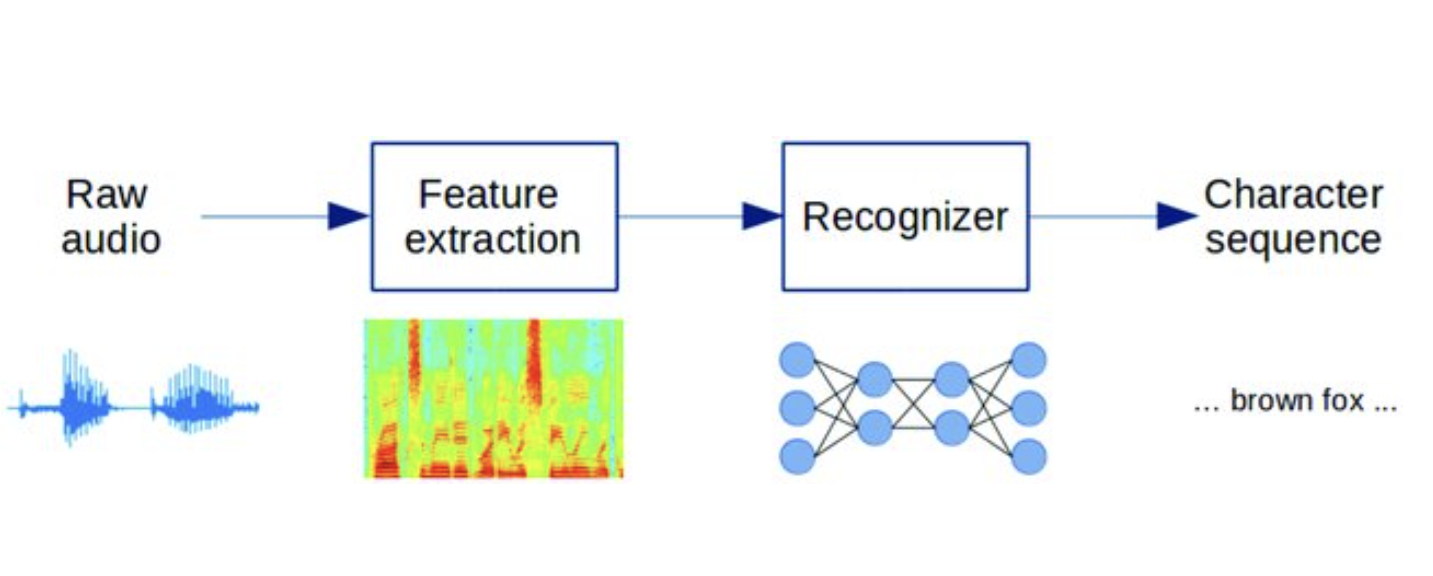}
  \caption{A basic End-to-End ASR architecture}
\end{figure}

However, despite the advances in Speech Recognition Neural Networks itself fail to solve 
problems without additional information. The problems faced by such techniques are mainly 
due to the unavailability of high resource data, this includes transcriptions, noise free datasets, 
specifications including Gender, Age, other information which can affect generalization 
capabilities. Other problems for End-to-End models can include environmental factors such as 
multiple-speakers, channel conditions, absence of Accent variations in the data, prosodical 
features, etc.

\subsection{Statistical Speech Recognition}
\begin{figure}[h!]
  \centering
  \includegraphics[]{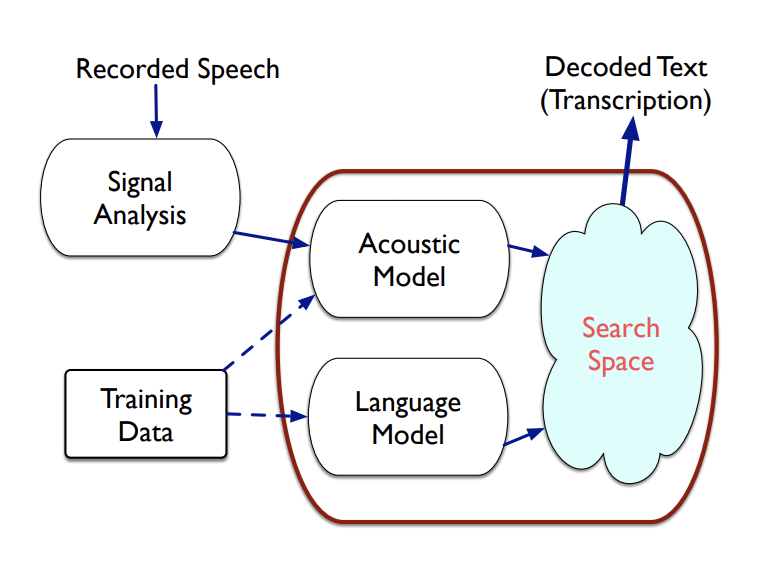}
  \caption{A GMM-HMM based ASR system}
  \label{fig: GMM-GM}
\end{figure}

Statistical speech recognition approaches Speech recognition problem in a different way than 
End-to-End models. It has different components, namely the Acoustic Model, Language 
Model, Dictionary which consists of Lexicons, Vocabulary; a Feature Extraction module and 
final all these modules are combined to probabilistically decode the final transcription. One of 
the advantages of this approach is the ability to train robust models using low-resource data. 
Also, since we have different modules, each of them can be tuned in a specific way given our 
problem domain for domain adaptations. This architecture also ensure high interpret ability 
since each result can be investigated as a series of probabilistic decisions during decoding and 
encoding phase.
We can formulate the Speech Recognition problem as below using Bayes Rule:\
\begin{center}
    $\hat{W}= \underset{W}{\mathrm{argmax}} P(X|W). P(W)/P(X)$
\end{center}
However, since we want the probability of $\hat{W}$ and the feature vector can be considered as a 
constant, so after simplifying:\
\begin{center}
    $\hat{W}= \underset{W}{\mathrm{argmax}} P(X|W). P(W)$
\end{center}
Where W is the list of sequences of words, X is the acoustic observations or feature vectors $\hat{W}$ the hypothesized word sequence. There are different parts to the Statistical model, these 
include the Acoustic Model, the Language Model, the Dictionaries. The following modules 
can be represented by the diagram below:

\begin{figure}[h!]
  \centering
  \includegraphics[]{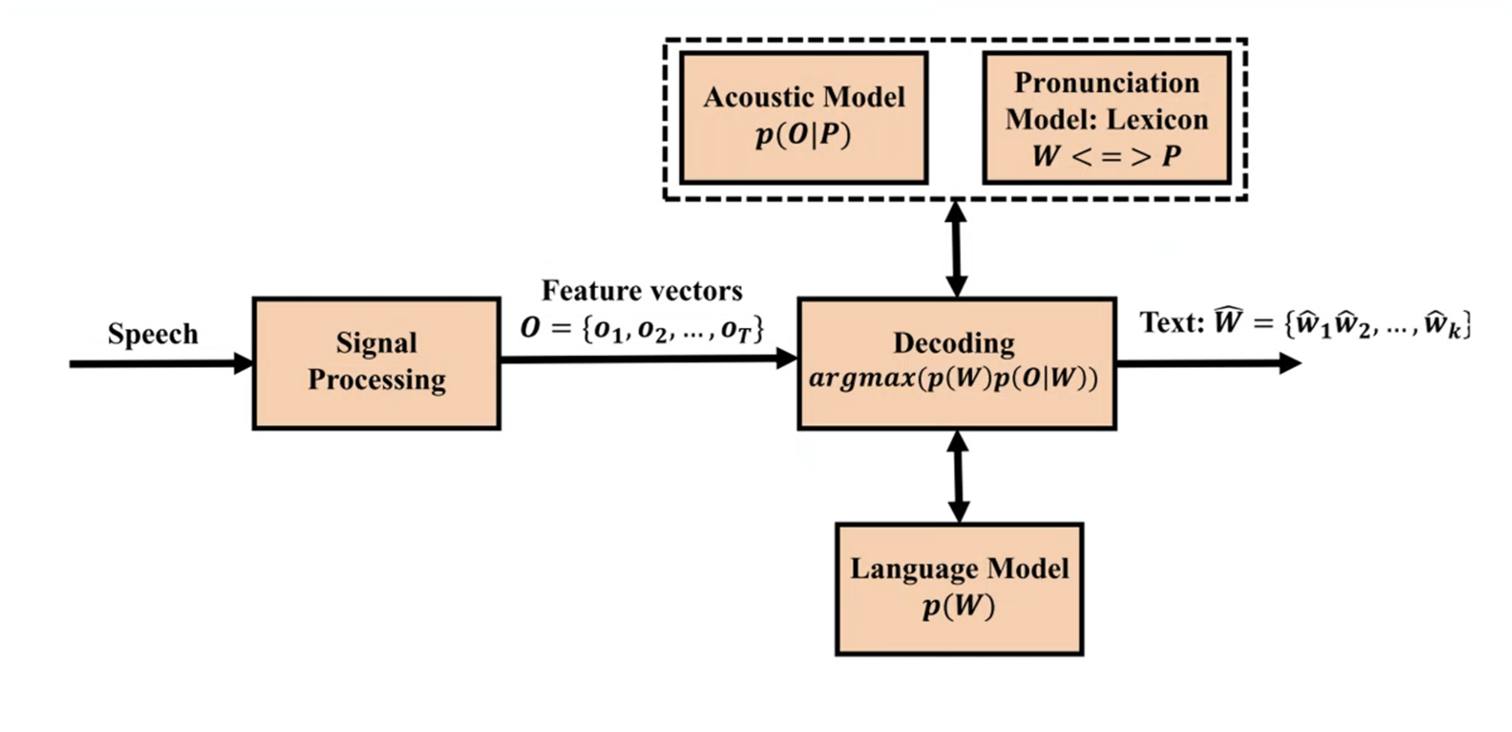}
  \caption{ Different blocks of an ASR system}
  \label{fig: Different blocks of an ASR system procedures.}
\end{figure}
\subsubsection{Acoustic Features}
To train our acoustic model responsible for giving the Observation Posteriors given our 
phonemes, we need to perform feature extraction on our raw signal. In ASR there are 
numerous different methods for feature extraction, some methods include training the raw 
signal \cite{ghahremani2016acoustic} using CNN and ReLU non-linearity, other features include Perceptual Linear
Prediction analysis (PLP) \cite{hermansky1990perceptual}, MFCC’s or the features commonly adapted for Speaker 
adaptation such as I-Vectors, X-Vectors \cite{snyder2018x}, D-vectors. In our setup, we chose an online approach which compute feature extraction on the fly. We used 39 Dim MFCC features 
with Delta and Double-Delta features. MFCC’s serve as the best baseline feature 
extraction method because of their ability to model the human cochlea. MFCC’s are also 
independent in nature thus, this makes it easy to model them since at each timestep $t$,the 
vectors are independent of each other i.e $P(O_t|O_{t+x})$ or
$P(O_t|O_{t-x})$
is False or simply put features do not have any correlations with each other.
\begin{figure}[h!]
  \centering
  \includegraphics[]{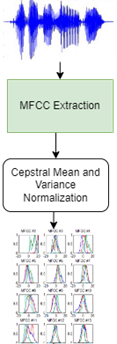}
  \caption{Normalized MFCC feature extraction}
  \label{fig:MFCC feat extraction}
\end{figure}
n our given use-case since we need our ASR to be robust, we require two types of 
feature-extraction methods. One for alignment stage, the other for computing the Deep 
Neural Network (DNN) posteriors. For the alignment stage, MFCC’s which have been 
normalized are sufficient. Another reason we use Mean and Variance features are to 
initialize the Gaussians of the GMM used per HMM state. This can be helpful for 
initialization during training HMMs. This makes them noise robust \cite{viikki1998cepstral} and are also used in the speaker-independent feature extraction stage described below:

\begin{figure}[h!]
  \centering
  \includegraphics[]{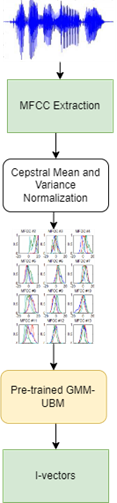}
  \caption{ I-vector feature extraction}
  \label{fig:I-Vector feat extraction}
\end{figure}

We extract speaker independent I-vectors using the normalized MFCC’s by feeding them to a GMM model trained universally on multiple speakers. Thus, it has learned to model in-speaker and between-speaker variabilities and create a compact low-dimensional vector. In our setup we used 40-dimension MFCC features with delta and double-delta features and 100 dimension I-vector features.

\pagebreak

\subsubsection{Acoustic Model}
This model is responsible for providing the probability estimates after seeing the 
Observation speech sequence. This model can be an GMM-HMM, DNN-HMM or 
TDNN-HMM, and other variants. The following model is trained on Speech features, to 
give the output probabilities. It is represented as $P(X|W)$ i.e., training the model to output
the probability of the speech sequence (frame level) given the transcript. Frames here refer 
to different parts taken from the Speech signal. Each signal is divided into frames and then 
these are processed by the acoustic model. This $P(X|W)$ usually the probability of 
observation sequence given the sequence of phonemes i.e. $P(X|P=W)$.

\subsubsection{Language Model}

The language model predicts the probability of the next word given the previous word. 
This can be modelled using N-gram approach, or by using other deep neural network based approaches such as RNN, LSTM, etc. This model is trained on a vocabulary of text 
to estimate the probability of the given word $P(W)$ .It is important to use N-gram models 
which use a variety of smoothing and interpolation methods for unseen sequences, which can the be modelled into a $G.fst$ from the language model.

\subsubsection{Pronunciation Dictionary}
To model the pronunciations, we need to create a dictionary of word and its respective 
pronunciations. i.e., grapheme and phoneme pairs. This is called the Lexicon. This lexicon 
is used to generate the list of phones, silence-phones, list of questions for decision tree based state modeling. We use the Silence phone as a way of representing the word 
boundaries. This is how we model the ASR system, using silences as a way of knowing 
when a particular word ends or if there are indeed silences due to stuttering. The
Librispeech Lexicon has been shown below:

\pagebreak

\begin{figure}[h!]
  \centering
  \includegraphics[]{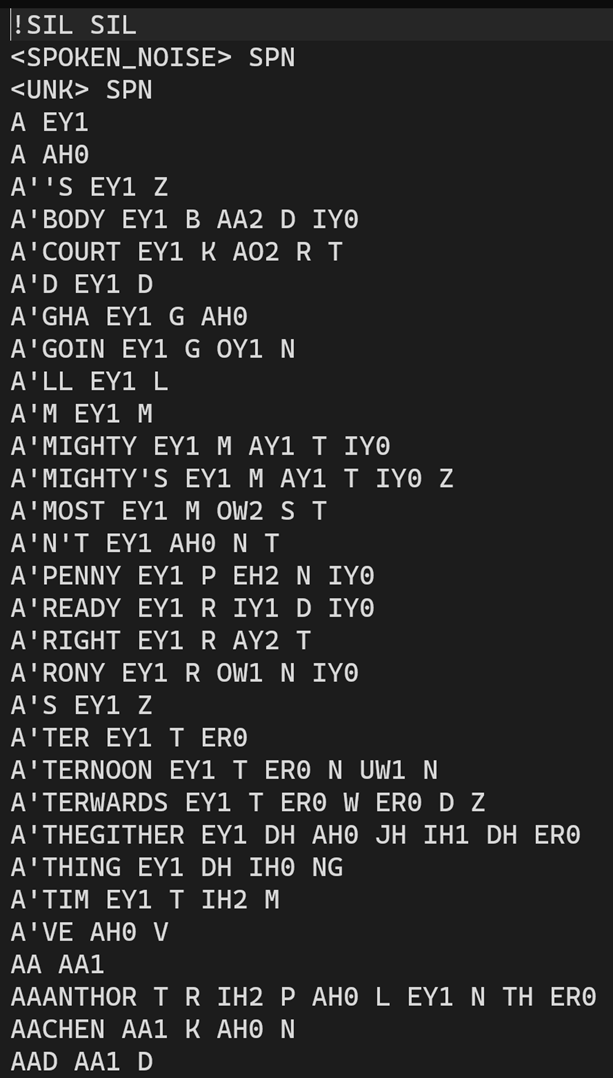}
  \caption{Librispeech lexicon}
  \label{fig:Librispeech lexicon cut out}
\end{figure}

The first 3 entries represent the silence, spoken noise, unknown word, and their respective
phonemic mappings. We describe the pipeline to create and modify the lexicon in later 
sections. Lexicons are extremely useful in variety of ASR related tasks. Such as given the 
grapheme and phoneme mappings one can train a Grapheme to Phoneme conversion 
module or using the concept of lexical compression create a corresponding data structure 
which essentially compressed the learned lexical compression. The same concept is used 
in the grapheme to phoneme tasks, since after “compression” it can essentially model new 
sequences to phonemes. Weighted Finite State Automatons (graph) are one such data 
structure used to represent the lexical information, other forms include the use of Prefix 
trees, Decision Trees, etc.

\pagebreak

\subsubsection{Decoding}

In this stage we find the likelihood of observing the given word sequence, given the 
Speech input. This process usually happens after the training has been done, and is usually 
performed at inference stage, although it does have its applications in the HMM training 
(forward-backward algorithm) and alignment stage. The Beam Search algorithm is used 
for alignments since they provide the best time vs cost complexity trade-off. The Figure 
below shows the training and decoding processes side-by-side. In the training steps, we 
require transcriptions, lexicons. In the decoding stage, we do not require any transcripts.
The algorithms for Decoding include Viterbi and Beam Search. We use the Beam Search 
algorithm due to the reduced time with which it can find a near optimal solution. The time complexity for Beam Search is reduced from $O(T.N^2)$ to $O(T.K^2)$, where is the beam-width which, $T$ is the time-steps and $N$ is all possible states. Here $K$ is always smaller than 
$N$. i.e., $K << N$ .This is especially useful while performing decoding over a huge graph 
constructed of multiple Weighted Finite State Automatons.

\begin{figure}[h!]
  \centering
  \includegraphics[]{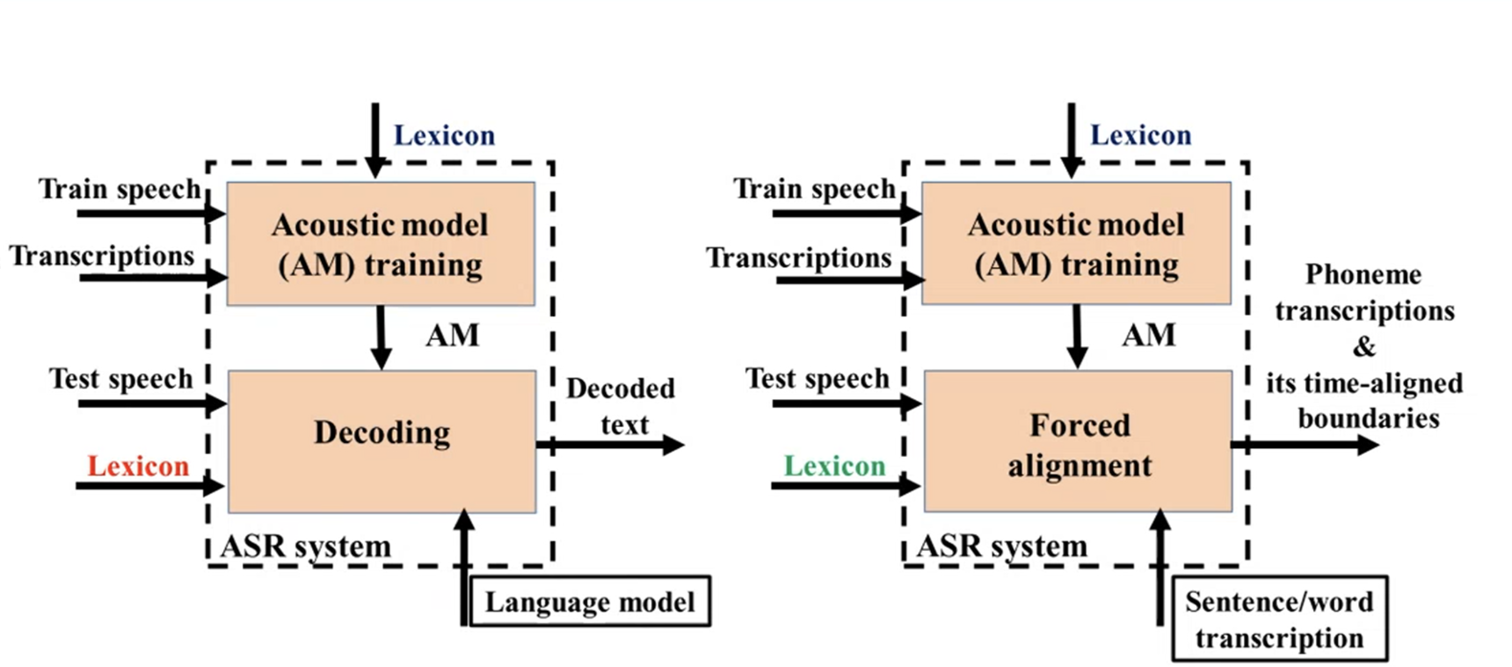}
  \caption{Decoding(Left) and Training(Right) procedures.}
  \label{fig:From Left to Right Decoding and training procedures.}
\end{figure}

In our task we are required to construct decoding graphs which in the form of Weighted
Finite State Transducers (WFST). We use the term WFST instead of WFSA since it is 
essentially an automation with inputs and outputs just like an actual transducer converting 
one quantity to another quantity. The decoding graph consists of 3 parts:

Here $H.fst$ represents a mapping from HMM/GMM states to context-dependent states.$C.fst$ represents a mapping from context-dependent phones or senones to the phones, and
$L.fst$ maps phones to the words. We map the $H$ to $C$ to $L$ using composition operations. 
Composition can be considered just as tensor dot products. \\
\begin{center}

\begin{table}[h!]
\begin{tabular}{ | m{5em} | m{5cm}| m{5cm} | } 
  \hline
  WFST & Input Sequence & Output Sequence \\ 
  \hline
  H & HMM States & CD Phones \\ 
  \hline
  C & CD Phones & Phones \\
  \hline
  L & Phone Sequence & Words \\ 
  \hline
  G & Words & Word sequence \\ 
  \hline
\end{tabular}
  \caption{HCLG Individual WFST components} 
\end{table}
\end{center}

\begin{figure}[h!]
  \centering
  \includegraphics[]{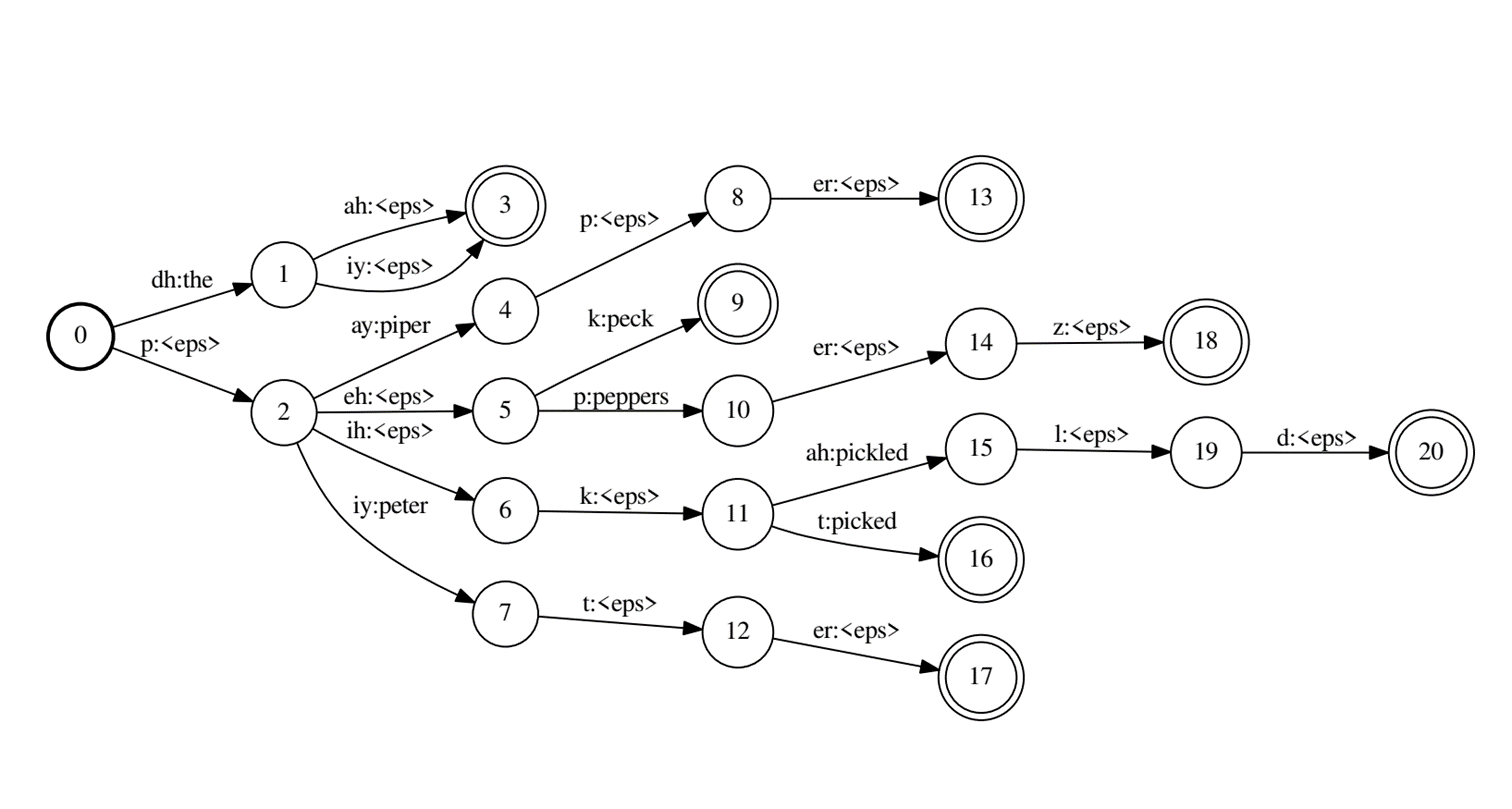}
  \caption{L.fst}
  \label{fig:L.fst}
\end{figure}

\begin{figure}[h!]
  \centering
  \includegraphics[]{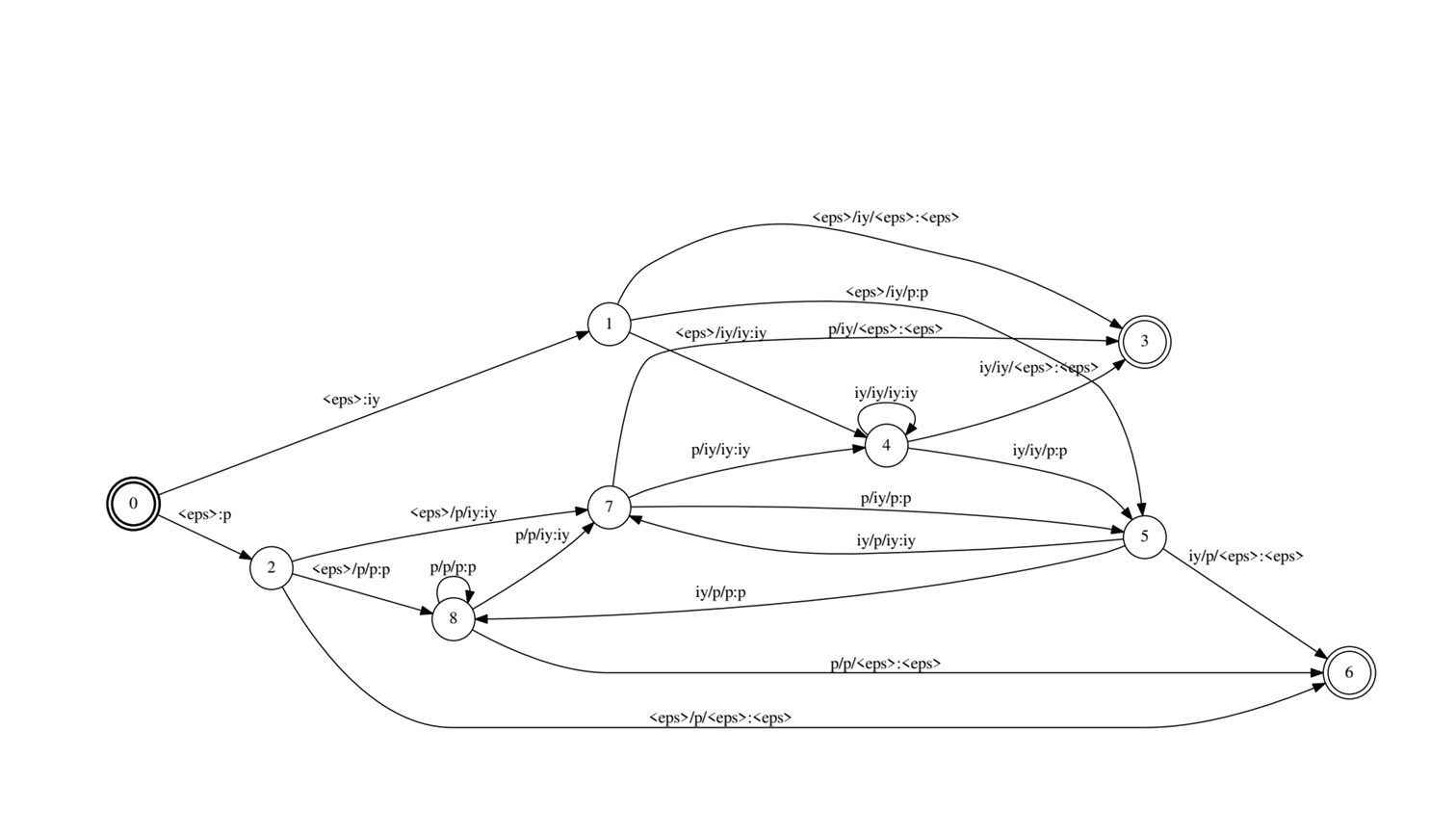}
  \caption{C.fst}
  \label{fig:C.fst}
\end{figure}

\begin{figure}[h!]
  \centering
  \includegraphics[]{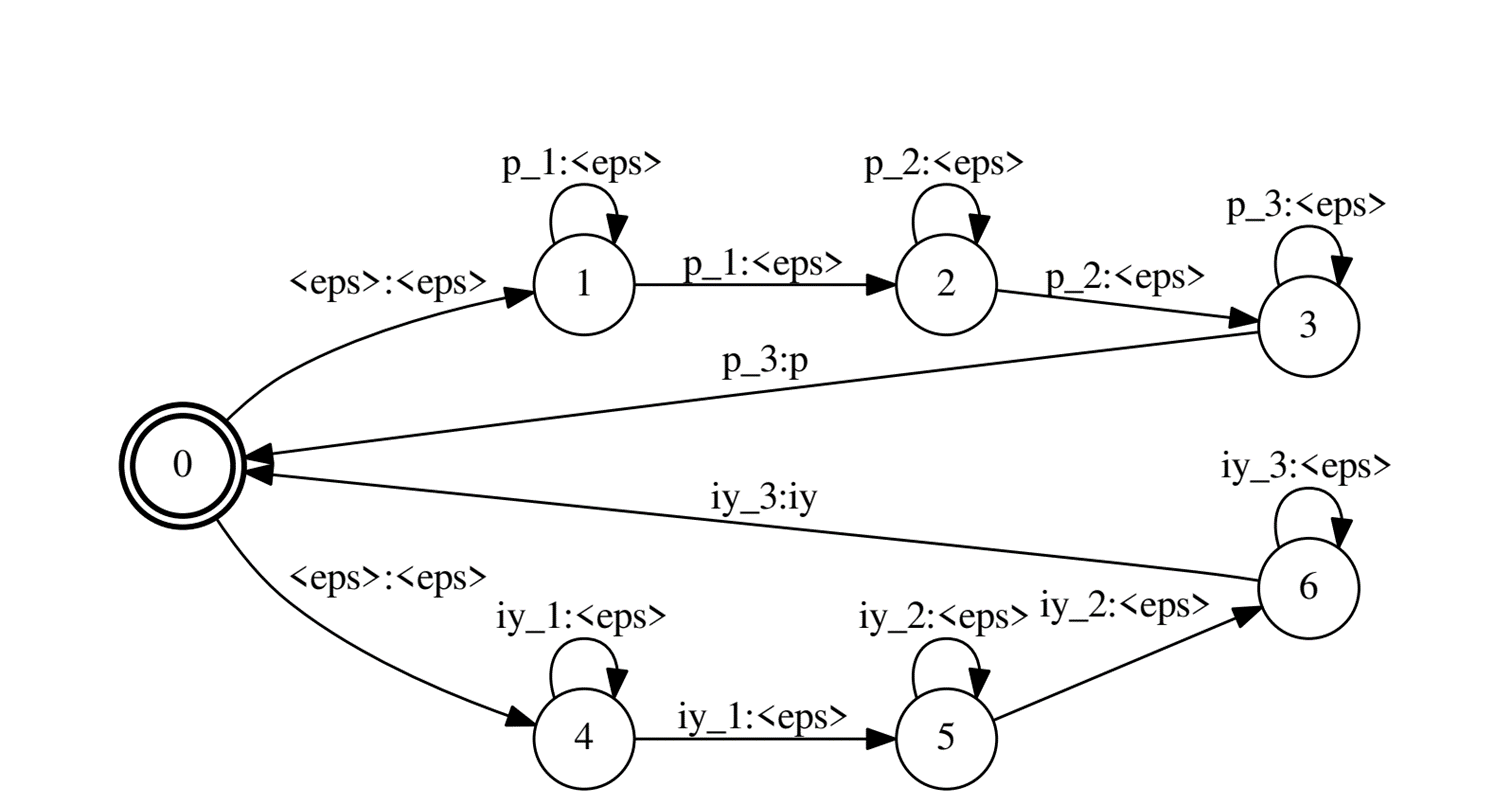}
  \caption{H.fst}
  \label{fig:H.fst}
\end{figure}

The inputs and outputs of the graph represent the corresponding input sequence and output sequence at each time point. These usually have a probability value concerned with the cost associated with each transition.

However, since WFST’s aren’t deterministic we apply operations to make them deterministic and need to apply minimization operation to reduce redundant paths. Another important thing required is the need for adding disambiguation symbol \#0 and self-loops in Kaldi to allow for composition of $L.fst$ with $G.fst$. Thus, the $L.fst$ with the disambiguation symbol is usually used during training such as the alignment graphs whereas the actual decoding is done using a $L.fst$ with added disambiguation symbols
The equation for the following is
represented below:

$
\mathbf{Standard \, ASR \, WFST \, Composition:} \newline 
HCLG = min(det(H \circ min(det(C  \circ min(det(L  \circ G))))))
$
\\[0.4 in]
$
\mathbf{Our \, WFST \, Composition:} \newline 
HCL = min(det(H  \circ min(det(C  \circ L))) 
$
\\

In standard ASR pipeline, we require hypothesis transcript as the output, however, in our
pipeline since we are only concerned with the phoneme sequences and already have the 
reference transcript at every speaker’s evaluation the need for a Language model and 
subsequent creation of Grammar $G.fst$ is eliminated.

\subsubsection{Alignment}
The alignment is essentially useful for aligning the source speech with our given 
transcript. This helps us train the acoustic models using Maximum Likelihood estimates. 
At each time-step of an iteration we get the alignments of the observed frames with the states of the HMM given an utterance. This is obtained while using forward-backward algorithm. Alignments are extremely helpful for us since we can automatically time align the transcript with the source speech. This reduces manual labelling effort. We 
obtain these alignments at phone level as well as word level.

\begin{figure}[h!]
  \centering
  \includegraphics[scale=1.1]{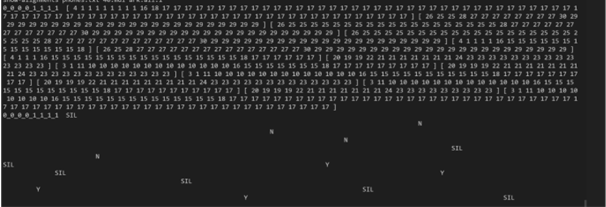}
  \caption{Raw alignments visualized}
  \label{fig:Raw alignments visualized}
\end{figure}

In the  Figure 11 we represent the raw alignments obtained after training a monophone HMM using GMM with Maximum Likelihood on the Yes/No corpus. Yes, grapheme is 
represented as Y and No as simply by N. These are the only phones present. The format of the above is the File-ID and the alignment vectors. The alignment vectors consist of each vector \[\] containing HMM state number, the next number represents the transition to the next state, transition to the same number represents the same state. Each input to the 
HMM state is essentially a frame, each vector [] represents a given phone segment. Within the segment are state number associated with the frame. State numbers not in sequence represent the \emph{senones} being captured for the given phoneme. The figure has to be read 
from the top starting with \emph SIL, after \emph SIL there is blank spaces representing same phone. After \emph SIL we get an N phoneme as an output, and then again blank space after which we 
get another N. Another phonetically time aligned alignment is shown below:

\begin{figure}[h!]
  \centering
  \includegraphics[]{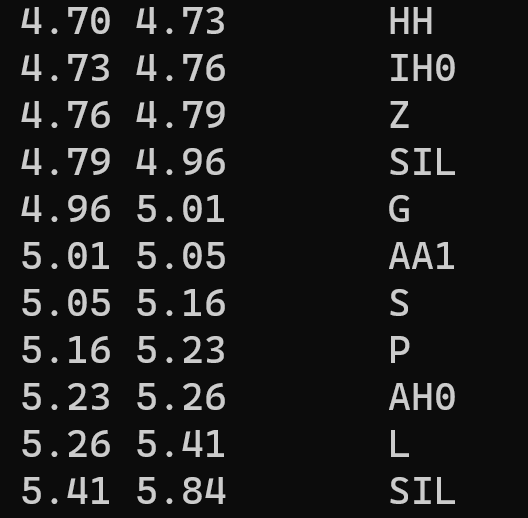}
  \caption{Phone level Conversation Time Mapping file}
  \label{fig:Conversation Time Mapping file}
\end{figure}

The given figure shows a .ctm file, this represents the alignment mappings with respect to 
time durations. The given utterance is taken from a part of an utterance of Librispeech and 
is read as “…HIS GOSEPL”.

\pagebreak


\section{HYBRID SPEECH RECOGNITION}

As discussed in the above Introduction, besides End-to-End and Statistical approaches another 
popular approach is the use of Hybrid Speech Recognition systems. These systems have a 
DNN model and an HMM model. Numerous hybrid approaches have been proposed. These methods combine the best of both approaches. We first describe the GMM-HMM approach to offer an introduction to how a DNN-HMM system would behave.

\subsection{GMM-HMM}

\begin{figure}[h!]
  \centering
  \includegraphics[]{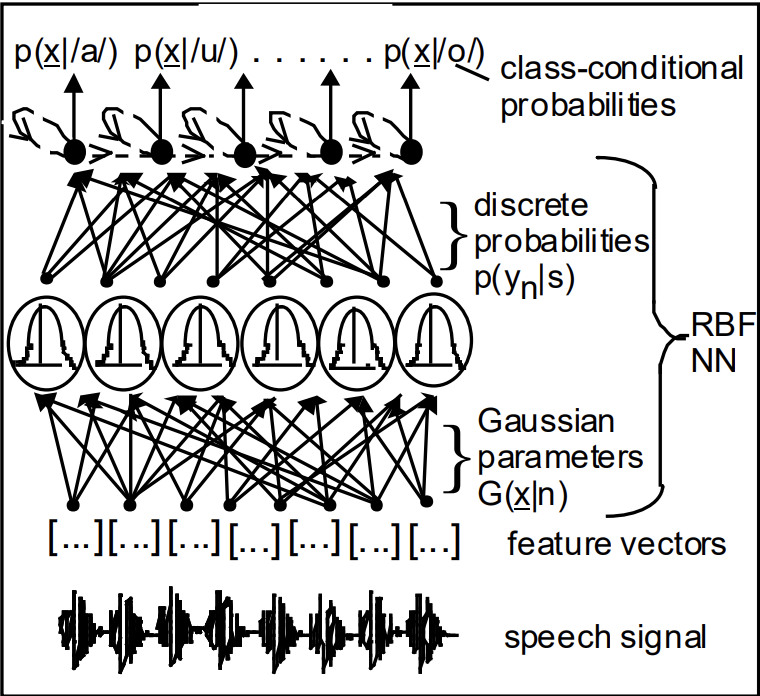}
  \caption{GMM-HMM model}
  \label{fig: GMM-HMM model}
\end{figure}

The GMM-HMM model is one of the earliest proposed models, although it is statistical in 
nature and not a hybrid it is important to understand GMM-HMM for the purpose of 
understanding Hybrid Systems
In this paradigm, each feature vector is modelled by a Gaussian. There are as many Gaussians 
as the number of HMM states or more. Each component of the GMM represents a Gaussian 
distribution. We use a 39 component Gaussian Mixture Model. Thus, we have 39 *44*3
mixtures. However, using the concept of tying or combining similar mixtures for similar 
phonetic states, we can reduce the number of states as well as capture phonetic context
dependencies. A Maximum Likelihood approach is used to estimate the training data. At each 
iteration the Likelihood increases, as we increase the Gaussian count. This approach is further 
improved by using a Decision Tree which asks a set of phonetic questions (E.g., Whether it is 
a stop? Or whether it is Left Vowel) and then assigns the phone states to a leaf node, thereby capture stress, tone related information as well as reducing the number of states since similar by clustering if they fall into the same leaf node.

\begin{figure}[h!]
  \centering
  \includegraphics[]{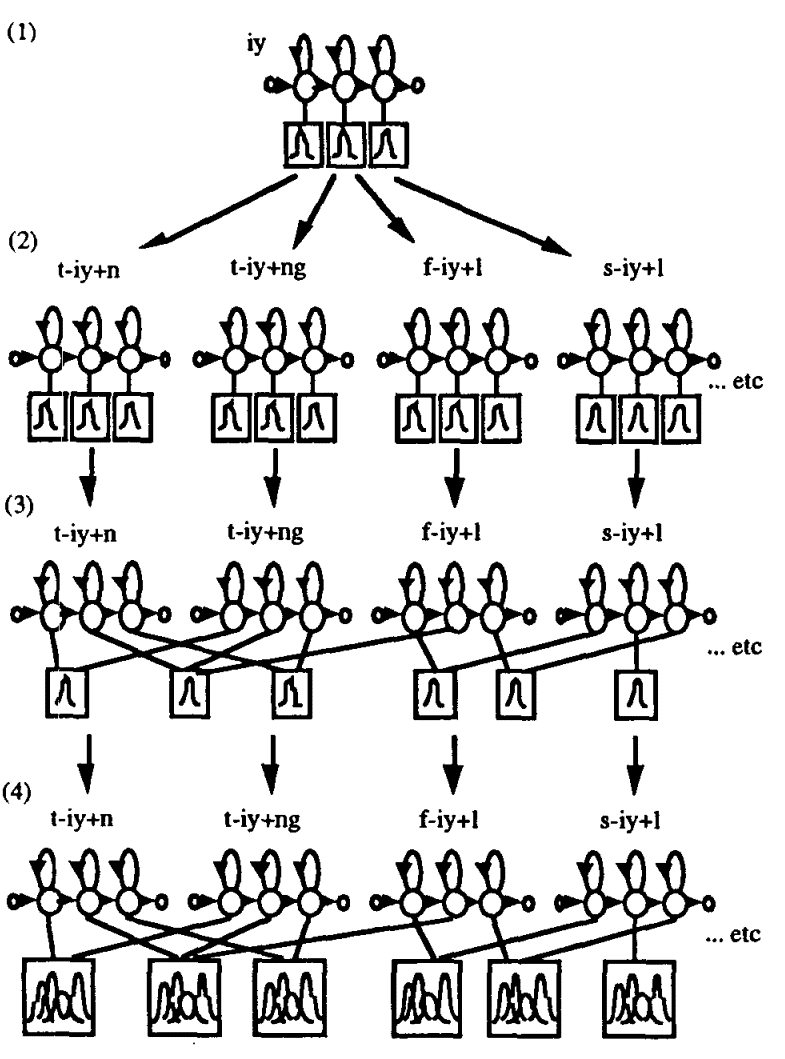}
  \caption{Tied state modelling in GMM-HMM}
  \label{fig: Tied state modelling in GMM-HMM}
\end{figure}

\begin{figure}[h!]
  \centering
  \includegraphics[]{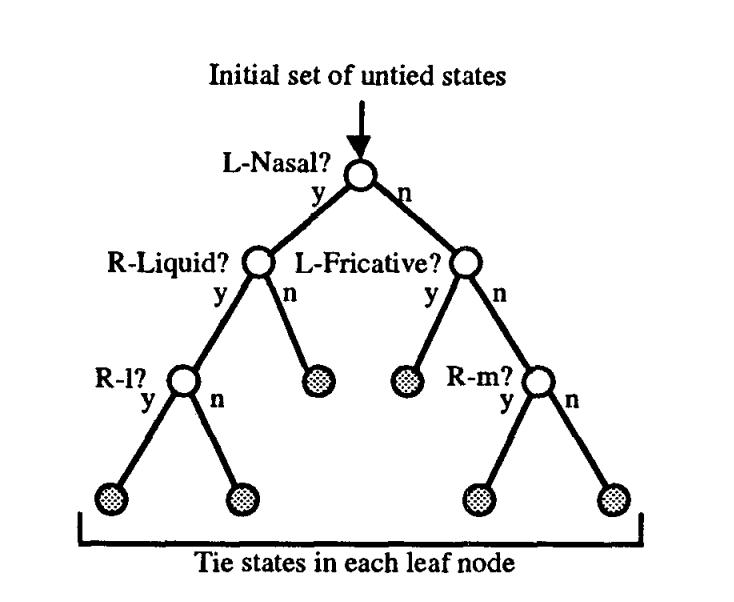}
  \caption{Tied state modelling using Phonetic Decision Tree}
  \label{fig: Tied state modelling using Phonetic Decision Tree}
\end{figure}

The advantages of this approach are usually related to unseen tri-phones. The states are tied 
[23] together by clustering the Gaussians of similar phones and the final tied states Gaussians 
are split to increase the number of mixtures. The GMM serves as an estimator for the HMM 
states.

\subsection{DNN-HMM}

In this method instead of using Gaussian Mixture Model as a classifier, we use a DNN \cite{young1994tree} as 
a classifier. The DNN’s final Softmax output layer, is a classifier with as many labels as the 
number of phones * the number of states per phone e.g. For 44 phones and 3 state HMM, it is 
44*3 labels to predict. This label assigns each acoustic frame to an HMM state.

\begin{figure}[h!]
  \centering
  \includegraphics[]{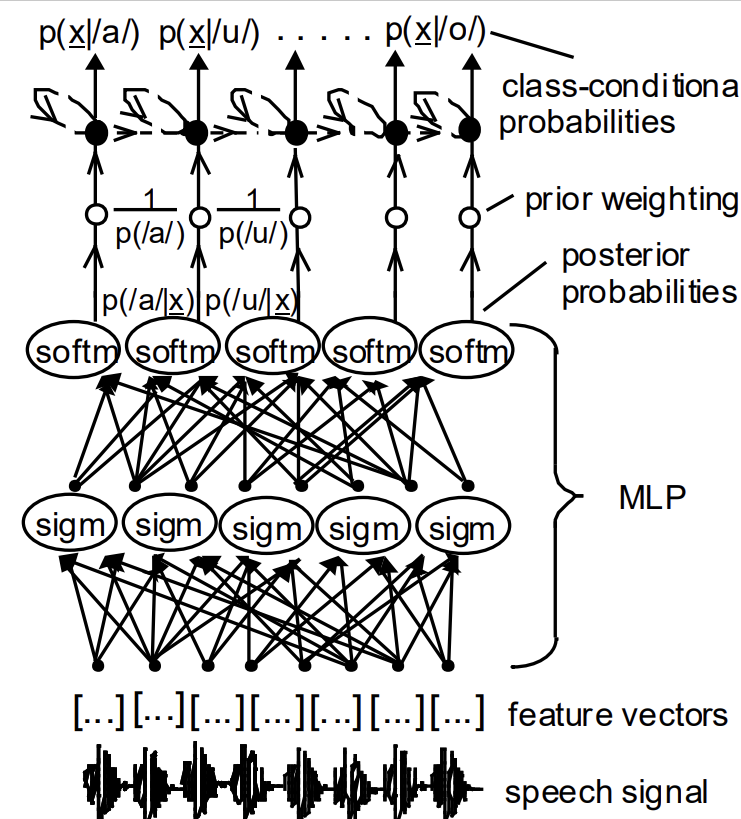}
  \caption{MLP based DNN-HMM model}
  \label{fig:MLP based DNN-HMM model}
\end{figure}

This assignment is in the form of posteriors $P(S|O)$. Using these state sequences, and the transition probability $P(S_2|S_1)$ between states we can get the $P(O|P)$ i.e $P(O|P=S_1..S_N)$ Here P is the Phone sequence. The neural network is discriminatively trained using the Cross Entropy 
Loss function on the predicted labels, which are the individual states to which the frame is assigned to. The advantages of training using Neural Networks are obvious. GMM’s require 
lot of parameters compared to simple Neural Networks such as RNN, also GMM assume the natural distribution of the data to be Gaussian which might not be true always. Moreover, 
GMM’s are trained in an unsupervised manner by maximizing the likelihood of the training data.

\pagebreak


\section{AUTOMATED PIPELINE FOR ACCENT 
QUANTIFICATION EVALUATION}
\subsection{Introduction}
Pronunciation evaluation is important for developing accent enriched speech recognition 
systems. An important step towards accent-based Speech modelling is also recognizing how a 
speakers accent affects the pronunciation, and how this could affect an ASR system. This is 
beneficial not only as a metric to recognize the pronunciation scores, but also can be used as a 
tool for automated scoring for assisting L2 learners.
\subsection{ Pronunciation Evaluation Methods}
Pronunciation evaluation methods have two types depending on the availability or in-availability of transcripts these are called Text dependent and Text independent pronunciation 
evaluation \cite{rigoll1998hybrid}. However, a common drawback of text independent methods includes the
absence of a verification agent with respect to the ground truth. There are several methods to
evaluate pronunciation of non-native speakers. Out of these the Goodness of Pronunciation 
(GoP) metric is one of the most used methods. This method proposed by \cite{neumeyer1996automatic} enables us to 
automatically score pronunciations based on log of posterior probabilities of phonemes 
normalized by the time. Further modifications proposed include the use of scaled log 
posteriors \cite{witt2000phone}, or the use of attention-based modelling to predict phoneme duration \cite{zhang2008automatic}, and 
transition factor between phoneme segments. Other methods include the use DNN based 
models instead of GMM’s for GoP \cite{shi2020context}. \cite{hu2015improved} explore how transfer learning approach could be \cite{huang2017transfer}
used in GoP task. However, most of these works are based on phoneme level modelling and 
do not consider senones \cite{hwang1992subphonetic} which allow parameter sharing thus reducing redundant states. 
Senones unlike tied states, magnify the contextual acoustic information, by not tying similar 
states but rather clustering them based on their distribution. We use a senone based approach 
as proposed by \cite{sudhakara2019improved} which also considers the state transition probabilities (STP) of HMM’s 
which are often discarded.

\subsection{GoP using Senones}
\begin{figure}[h!]
  \centering
  \includegraphics[]{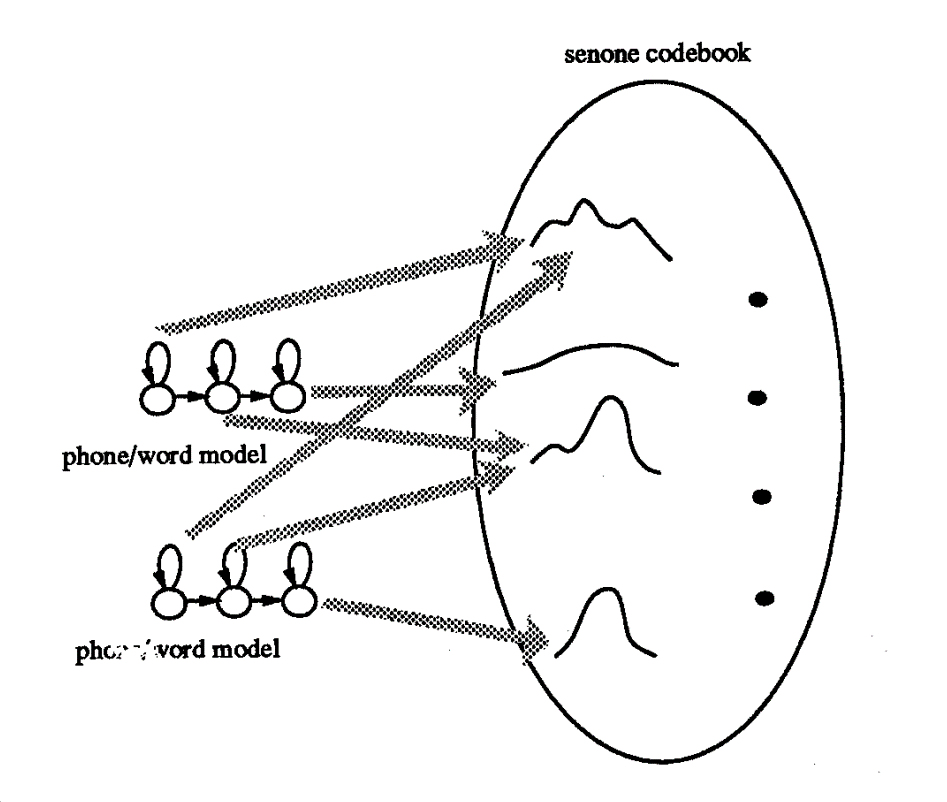}
  \caption{Position dependent phonemes clustered into Senones}
  \label{fig:MLP based DNN-HMM model}
\end{figure}
As discussed above we use the senone based approach. We discuss the details of the senone 
based approach:
\begin{center}
    \Large \textbf{$GoP(p)=\frac{1}{T} | \log{\mathcal{P}(p|O)} |$}
\end{center}

\begin{center}
  \includegraphics[]{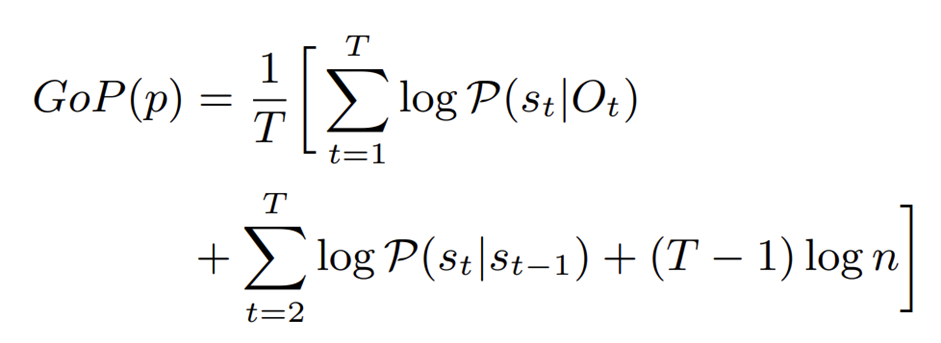}
\end{center}

Here the first term represents the probability of belonging to $s_t$ given the Observation 
sequence. The second term represents the senone transition probabilities added with the log of 
number of senones. In the sections below we describe the methods in which we calculate the
necessaries for the given equation and build a pipeline around it

\subsection{Overview of Pipeline}

In this section we describe the pipeline that we created for automatic computation of 
Goodness of Pronunciation scores. Another reason to choose the improved GoP metric was 
it’s Open-Source availability and ease of use. The metric also doesn’t utilize any heavy 
models based on attention such as transformer layers or stacked RNNs. This make it 
extremely useful for creating Online pipelines, which would have been difficult due to latency 
of other large models. This is important since it shows higher correlation to human evaluators. 
We describe the various stages of our pipeline below:\\

\begin{figure}[h!]
  \centering
  \includegraphics[]{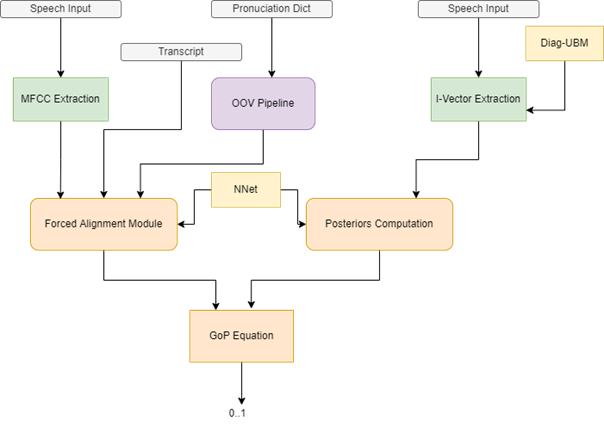}
  \caption{GoP pipeline overview}
  \label{fig:GoP pipeline overview}
\end{figure}

The complete pipeline overview has been shown above. The MFCC and I-Vector Extraction 
modules have been already described in the Acoustic Features section. The Forced Alignment 
module is again described in the previous section. A slight change however is the use of DNN-HMM based alignment rather than GMM-HMM. The I-Vectors are extracted using a 
pre-trained speaker independent GMM with a diagonal covariance matrix (only diagonal has standard deviation information, rest are zeros or sparse) to get speaker independent information. The Nnet is a neural network model, an MLP with 25 layers, with outputs being the (senones) tied states of a Context Dependent (Triphone) HMM. In the posterior computation stage, we use a pre-trained DNN-HMM model. For us to automatically evaluate 
the pronunciation this model was trained on native speech data i.e., Librispeech (960 hours) and Fisher English corpus.This ensures the posterior estimates of the model are with respect 
to native speech and thus it is useful when non-native speech is used as input to the model.Non-native speech will have a characteristic accent, which will lead to changes in the 
probability estimates of the senones. This can be considered as removing the native speaker who is the human evaluator and replacing it with a natively trained DNN-HMM.

\subsection{Challenges}
Before we explain the pipeline, itself we need to understand what challenges one faces while 
incorporating the GoP equation into our pipeline.

\subsubsection{Technical Challenges}
As with any code, code breaks and often becomes un-maintainable. Our challenges in the 
technical domain involved fixing backward compatibility issues, in our experiments we use a 
robust Grapheme to Phoneme model called SequiturG2P which models the grapheme to 
phoneme problem using graphone units and M-gram approach with smoothing and 
interpolation. We used this to model the lexicon (describe in later sections). The initial 
challenge with G2P was the issue with backwards compatibility which was fixed by the 
author of this report. This enabled backwards compatibility across Python2.
While preparing the dictionary related files for the given database, there were issues prevalent
with respect to single-threaded performance. This would render the entire pipeline unusable 
and for other users to utilize a single thread. The issue would cause a bug, leading to 
previously run multiple thread intermediates to be used again or fail in single thread mode.
This again was fixed by the author of this report. Now, we describe one of the biggest issues faced by the authors, which is the problem of Out of Vocabulary occurrences. OOV is a big issue in our pipeline since any OOV occurrence would be mapped entirely as a \emph{SPN} phoneme at testing time. This isn’t ideal when we are trying to build an automated system for phoneme-based pronunciation scoring. On further investigation we found the reason for the issue was with respect to data set(Librispeech,etc), to eliminate some words so as to not expand the Lexicon size. Despite the innocence and factually supported decision, it is not ideal for such cases where accurate phonemic alignments are required for scoring. The pipeline is described in Sections below.

\subsubsection{Out of Vocabulary at Testing time Challenges}

The OOV pipeline essentially is an automatic pipeline which can generate Lexicons for words 
which are OOV for our given databases in question. In our case we used the CMUDict and 
Librispeech database. Any word does not present in either of the two is mapped as SPN
during the alignment outputs. Thus, we need to formulate an automated procedure to get these 
phoneme mappings. This is essentially the method used by us to generate the Lexicon. 

\begin{enumerate}
    \item Compare the vocabulary entries with CMUDict lexicon for corresponding phoneme 
mapping.
\item Compare the vocabulary entries in (original Librispeech) lexicon.
\item If all words are found, we stop and get list of all mapping in a lexicon file.
\item If words are not found in either database, we automatically generate them.
\item Pass the OOV words to Grapheme to Phoneme model.
\item Append the automatically generated lexicon to lexicon entries which are not OOV.
\item Sort and combine these two into a final lexicon containing all lexicon entries including 
OOV.
\end{enumerate}

\pagebreak
\begin{figure}[h!]
  \centering
  \includegraphics[]{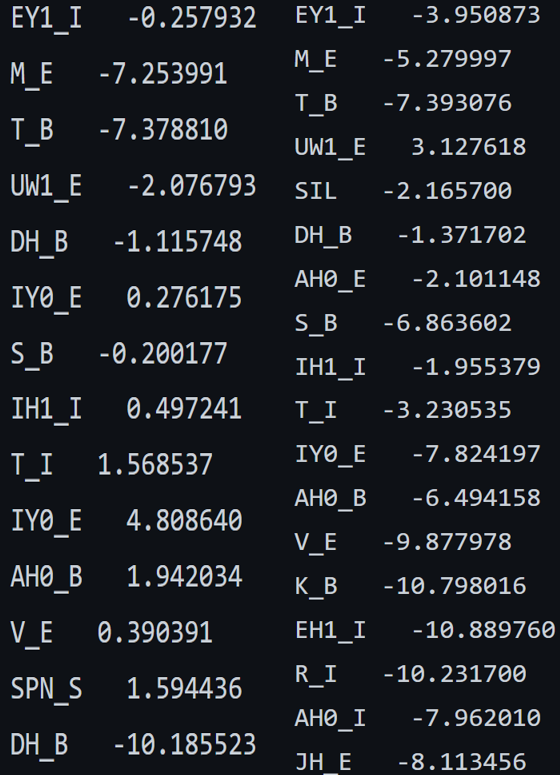}
  \caption{GoP outputs with and without OOV (Side by side)}
  \label{fig:GoP outputs with and without OOV}
\end{figure}

This makes sure that the Lexicon we use in our pipelines models these OOV occurrences and doesn’t lead to any errors downstream in Lexicon Transducer compilation and subsequent 
forced alignment. The final output of our system is a phonetically aligned transcript with GoP probabilities per phoneme of the utterance.

\paragraph{Grapheme to Phoneme Conversion}

For Grapheme to Phoneme conversion, we used Sequitur G2P model. This model uses the 
concepts of units called graphones which are essentially a unit of grapheme and phoneme 
together. We use the grapheme to phoneme convertor called Sequitur G2P [17] in essence it 
finds the Grapheme- Phoneme mappings from the training data(lexicon). These mapping for a 
particular sequence (of a word) are called alignments or segmentations. A graph of all 
possible segmentations is constructed and the maximum likelihood $\mathcal{L}(g,p,q)$of the training data (grapheme, phoneme sequence pair) is maximized given our hypothetical 
segmentation which we are trying to obtain. Once we obtain such a hypothetical alignment, we can update a M-gram model. The M-gram model is essentially a sequence of graphones 
instead of words. We use a pre-trained 6-gram model. Once the likelihood function saturates, we can use those obtained segmentations to update the M-gram model. However, M-grams
are prone to OOV, so an integrated smoothing and interpolation approach is used. Here, we 
estimate discount parameters using Maximum Likelihood objective using Expectation 
Maximization algorithm again on a held-out set not seen during training, and obtain the 
discount values $d_M$.

For each M-gram. This integrated approach also helps us estimate the back-off distribution of $p_{M-1}(q)$ the lower order M-gram. This basically assigns some probability mass by 
redistributing it from the graphone occurs which have extremely high probability mass using 
discount parameters and the backoff-distribution when necessary.
Once the discount parameters are known, and the N-gram updated. A decoding graph is 
constructed from the segmentations in the form of a Finite State Automata, where each (node) 
grapheme input corresponds to an output phoneme associated with the previous graphone 
history. The weight of each arc has a probability assigned to it which depends on the graphone 
sequence’s M-gram probability calculated. The input to this graph will be a sequence of 
letters from a grapheme, and after doing decoding on the graph a graphone sequence is 
obtained. Using this graphone sequence we can get the phoneme sequence as well. This is our 
most likely pronunciation. Several methods for decoding can be used. The simplest being 
First Best Search giving the best path. However, multiple pronunciations (phoneme 
sequences) are also possible, so after Beam Search through the graph (FSA) we can collect 
these alternative paths and store them in an array/stack like structure. We can create another 
graph after getting paths which may have repeated nodes, from this graph we can again do 
decoding by A* search and get the new rescored values and chose the top-n or the one with 
best cost.

\subsection{Pipeline Methodology}

As discussed in the sections above, it is evident that building such a workflow requires a 
meticulously designed pipeline to enable the automatic score computation of millions of 
Speech files. Speech files could be in \emph{.wav} or \emph{.flac} or \emph{.mp3} format.We consider these 3 file 
formats at present. As discussed in Sections above, the above the pipeline is not well-defined with respect to the cases of Out of Vocabulary occurrences and incorporating this pipeline 
into our system is challenging. However, we devise 3 simple pipelines which are able to integrate the OOV component into our existing architecture.
\pagebreak
\subsection{Offline}
In this mode the Out of Vocabulary problems are fixed before the execution. i.e., before any 
inputs are given to the pipeline. This means we pre-compile a set of words that resulted or 
could result in OOV for a given database. This means that any OOV other than those already 
present might fail and the process stops. However, it works well in the assumption that no 
OOV occurs beyond these additions. A very crude assumption which saves executing the 
OOV pipeline during execution but has its flaws. We describe method for the Offline mode:

\begin{figure}[h!]
  \centering
  \includegraphics[]{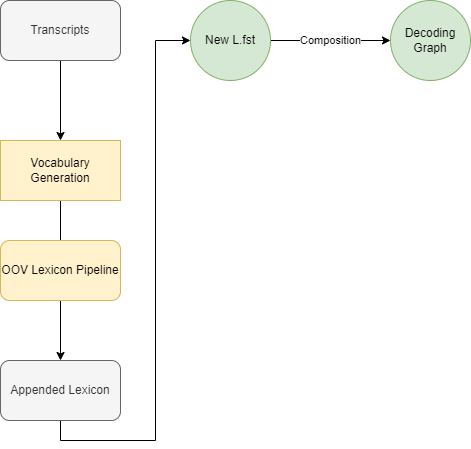}
  \caption{Offline Pipeline}
  \label{fig:Offline Pipeline}
\end{figure}
\pagebreak
\begin{enumerate}
    \item Collect all Transcripts for Librispeech database. Automatically generate a new vocabulary.
    \item From the collected transcripts. Generate a new Lexicon using the Lexicon generation
    \item Generate Lexicons from OOV using the pipeline and keep appending it.
    \item Get the final Lexicon with OOV lexicon entries. Get list of silence, non-silence phones, 
other files from the lexicon. Compile the Lexicon L.fst from the Lexicon and place in language model folder.
\item  This folder containing the lexicon, phones, L.fst is used every time
for Decoding Graph Creation. 
pipeline.
\end{enumerate}

\subsection{Online}

In this mode the Out of Vocabulary problems are fixed during the execution, compared to 
Offline where we pre-compile the Decoding Graph from a pre-appended Lexicon with OOV
entries which are known to be encountered. Thus, this method results in automatic creation of 
Lexicon, the Decoding Graph Composition, and the final output with respect to the given 
utterance. The Online mode can have 3 variants; however, we settle with these methods 
because our best method was incompatible with Kaldi’s standard architectures.

\subsubsection{Vocabulary Expansion Based}

\begin{figure}[h!]
  \centering
  \includegraphics[]{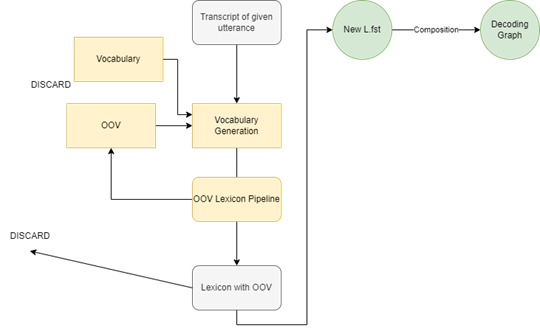}
  \caption{Online Vocabulary Expansiion  based Pipeline}
  \label{fig:Offline Pipeline}
\end{figure}

\begin{enumerate}
\item Get Transcript for current utterance. 
\item Find words from current transcript not in any of the 
databases available. 
\item Add those OOV words to Librispeech-vocabulary. 
item Regenerate lexicon from appended vocabulary, using Lexicon generation pipeline. 
\item Generate the list of 
silence, non-silence phones, extra-questions for Decision Tree clustering automatically. 
\item Compile Lexicon Transducer L.fst from new lexicon. Create the language model folder 
and use it for computing HCL.fst Decoding graphs. 
\item Repeat Steps each time OOV occurs.
\end{enumerate}

\subsubsection{Lexicon Expansion Based}

\begin{figure}[h!]
  \centering
  \includegraphics[]{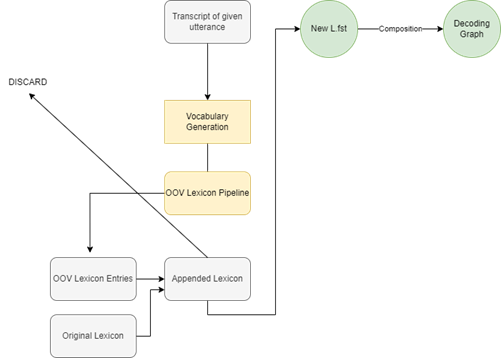}
  \caption{Online Lexicon Expansion  based Pipeline}
  \label{fig:Offline Pipeline}
\end{figure}

\begin{enumerate}
\item Get Transcript for current utterance. 
\item Find words from current transcript not in any of the 
databases available. 
\item Add those OOV words to Librispeech-vocabulary. 
\item Regenerate lexicon from appended vocabulary, using Lexicon generation pipeline. 
\item Generate the list of 
silence, non-silence phones, extra-questions for Decision Tree clustering automatically. 
\item Compile Lexicon Transducer L.fst from new lexicon. Create the language model folder and use it for computing HCL.fst Decoding graphs. Repeat Steps each time OOV occurs.
\end{enumerate}

\subsection{Hybrid}

\subsubsection{Lexicon Expansion Based}

\begin{figure}[h!]
  \centering
  \includegraphics[]{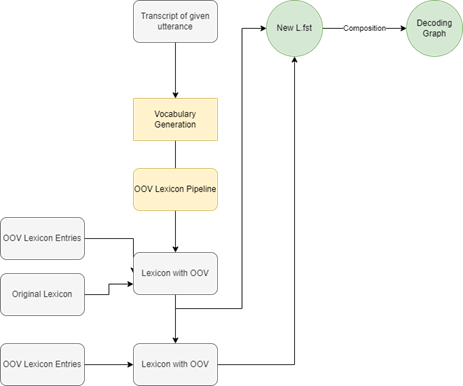}
  \caption{Hybrid Approach}
  \label{fig:Offline Pipeline}
\end{figure}

In Offline mode the lexicon stays constant, in Online mode the Lexicon is appended but 
changes are never preserved in them. Thus, the change is only present for each utterance and 
then discarded. This means that the entire process must repeat again and again despite adding
for the given utterance the changes are not preserved in memory. We fix this by proposing a 
hybrid pipeline. In this we follow the same steps as the Online mode but make a conscious 
decision to preserve the Lexicon. This successfully solves the issue of repeated Online 
computations and reduces the number of Online runs to one for each unique OOV 
encountered in the utterance. This pays us dividends by reducing computational cost. The 
only disadvantage of preserving the Lexicon is the growing size of the Lexicon. We discuss 
these challenges in Section 5.7. The Hybrid process is described below:

\begin{enumerate}
    \item Get Transcript for current utterance. Find words from current transcript not in any of the 
databases available.
    \item Generate lexicon for the OOV vocabulary words.
    \item  Regenerate lexicon 
from appended vocabulary, using Lexicon generation pipeline. Generate the list of silence, 
non-silence phones, extra-questions for Decision Tree clustering automatically.
\item Compile
Lexicon Transducer L.fst from new lexicon. Create the language model folder and use it for 
computing $HCL.fst$ Decoding graphs.
\item Repeat Steps 1-7 each time OOV occurs, but do not 
discard the Lexicon.

\end{enumerate}
\pagebreak

\section{SUMMARY AND DISCUSSION}
We describe three methods by which we can create an optimized pipeline for automatic GoP 
score computation. The above methods integrate the OOV pipeline by proposing different 
methods for optimizing the cost associated with the space and time complexity. We also provide utilities to compute Phoneme and Word level alignments,  Phoneme posteriors which can be used by researchers in the future for other downstream tasks associated with GoP task, or other tasks. Our method is somewhat of an Hybrid between On-the-fly composition and Fully Composed static WFSTs for decoding since the $L.fst$ is re-constructed as and when required based on the input utterance.\cite{hori_speech_2013} A major problem noted with this method is the presence of dead-states during the decoding phase.

\begin{figure}[h!]
  \centering
  \includegraphics[]{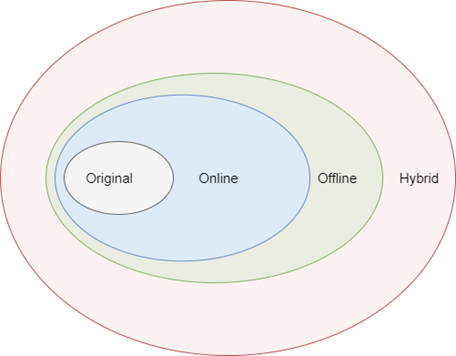}
  \caption{The vocabulary space denoted as sets.}
  \label{fig:The vocabulary space denoted as sets}
\end{figure}

In this Section we discuss about the consequences of these methods on the Lexicon and 
Lexicon Transducer. The Offline Lexicon can be considered as a \emph{static} lexicon, and the others
can be considered as a \emph {dynamic} lexicon. The Offline Lexicon $L_{Offline}$ is the biggest 
Lexicon at experiment start.Let us consider the initial Lexicon(without any additions) as  $L_{0}$  which does not contain any additional OOV entries.  The 
other Lexicons $L_{Online}$ and $L_{Hybrid}$ are equal to $L_{0}$ before experiment start.However, the $L_{Online}$ size can never expand to be that of $L_{Offline}$  since changes to it are discarded after each addition of OOV from a new utterance. After 1
addition provided the same utterance $L_{Hybrid}$ and $L_{Online}$ can be equal as
\begin{center}
    $L_{Online} = L_{Online} = L_{0} . L_{1-entry}$
\end{center}

where $L_{1-entry}$ denotes the lexicon FST of the given addition.

\begin{center}
    $L_{Offline} = L_{Hybrid} = L_{0} . L_{1-entry}....L_{N-entry}$
\end{center}
Moreover, a situation can occur such that $L_{Hybrid} >> L_{Offline}$ large number of 
inputs cause the Hybrid Lexicon to increase in size beyond the Offline lexicon entries. . This 
31
can occur in real-time since OOV size would be unlimited beyond the scope of the OOV 
present in the database lexicon , derived from its transcriptions. The following is summarized
in the above figure, which represents the given equation visually in the form of the space of
vocabulary as sets. The universal set U or an hypothetical lexicon FST $L_U$  containing all possible 
words known to mankind  $L_U \ne L_{Hybrid}$  due to practical limits of space complexity.

\pagebreak

\section{CONCLUSION AND SCOPE OF FUTURE WORK}

In the following project we propose the creation of a pipeline for the use of a Pronunciation evaluation. We further 
highlight the various issues one faces while formulating a pipeline of automatic score 
computation. Different methods for the pipeline have been proposed. We propose pipelines to incorporate OOV problem at  during Goodness of Pronunciation evaluation/testing time when scores are being computed.

\subsection{Conclusion}
Traditional ASR systems can be trained to model accent at various stages of the ASR 
pipeline. Out of Vocabulary can be an issue for ASR systems and solving them can be critical for systems which depend on phoneme or word alignments for downstream tasks.Language modelling using WFST representation can be an technique considered for vocabulary compression. Expanding the Lexicon can lead to state space explosion of the corresponding WFSTs.Thus, there exists a clear trade-off between the size and decoding accuracy.  Composing reconstructed(re-compiled) WFSTs can be done in a variety of ways, each having it's own tradeoff for time vs space complexity.

\subsection{Future Work}
There are numerous unexplored ideas that arise from our given work. We discuss some of these ideas that could arise as a result or adjacent to our research.
\begin{itemize}
\item Further optimization of $L.fst$ to improve decoding.
\item Methods involving On-the-fly Composition of the $L.fst$ and On-the-fly Rescoring can be easily alleviate some of the concerns during Online computation.
\item Explore how the Phoneme posteriors and GoP scores could be used as speech features (vectors as inputs) to Neural Network architectures such as Wav2Vec, HuBERT ,etc for improving Accented Speech Recognition and potentially for Prosody related tasks.
\item Exploring the generalization capabilities of Neural Network architectures such as 
RNN and Transformers for posterior computation in GoP.
\item Future systems which could potentially utilize LLM's along with Lexicons for improve GoP.

\end{itemize}
\pagebreak
\bibliography{Library_for_Automatic_References.bib} 

\end{document}